\documentclass{article}
\usepackage{spconf,amsmath,graphicx}
\usepackage{booktabs}
\usepackage{array}
\usepackage{enumitem}
\usepackage{url}
\usepackage{tikz}
\usetikzlibrary{arrows.meta}
\setlist{nosep, leftmargin=14pt}

\makeatletter
\let\@sect@spconf\@sect
\def\@sectunnum#1#2#3#4#5#6[#7]#8{%
  \begingroup
    \ifnum #2=1\bf\centering
      {\interlinepenalty\@M \uppercase{#8}\par}%
    \else\ifnum #2=2\bf
      \noindent{\interlinepenalty\@M #8\par}%
    \else\it
      \noindent{\interlinepenalty\@M #8\par}%
    \fi\fi
  \endgroup
  \@tempskipa #5\relax
  \@xsect{\@tempskipa}}
\renewcommand{\thebibliography}[1]{%
  \let\@sect\@sectunnum
  \section{References}%
  \let\@sect\@sect@spconf
  \list{[\arabic{enumi}]}{\settowidth\labelwidth{[#1]}\leftmargin\labelwidth
    \advance\leftmargin\labelsep\usecounter{enumi}}%
  \def\newblock{\hskip .11em plus .33em minus .07em}%
  \sloppy\clubpenalty4000\widowpenalty4000\sfcode`\.=1000\relax}
\makeatother

\title{NucShare: Lightweight Nuclei Instance Segmentation with Foundation-Model Accuracy}

\name{Wenyan Li, Xiangyang Jia\sthanks{Corresponding author.}, Wenlong Zhong, Xu Chen}
\address{School of Computer Science,\\
Wuhan University, Wuhan, China}

\begin{document}
\maketitle

\begin{abstract}
Nuclei instance segmentation is fundamental to quantitative pathology. Yet high-accuracy methods
mostly build on large vision transformers (ViTs) or heavy convolutional backbones, whose parameter count and
compute cost hinder deployment in clinical and edge settings. We present NucShare, a lightweight model whose
shared decoder handles nuclear pixel segmentation, horizontal--vertical distance, and classification through
learnable transposed convolutions for upsampling and three lightweight heads, using 37\% of independent
decoders' parameters with no loss in accuracy. Output-level distillation then transfers knowledge from NucShare-T (UNI2-h)
to NucShare-S (ConvNeXt V2-Tiny), reducing 690.5M parameters to 34.7M. On PanNuke they reach mPQ 0.524 and
0.519, above the prior best CFR-SAM-H at 0.511, and NucShare-S surpasses HoVer-Net and Mask R-CNN by 5.6
and 15.0 points; on zero-shot MoNuSeg it attains bPQ 0.668 against CFR-SAM-H's 0.662. Code:
\url{https://github.com/L-iMeng-L/Newidea}.
\end{abstract}

\begin{keywords}
nuclei instance segmentation, knowledge distillation, lightweight model, efficient inference, digital pathology
\end{keywords}

\section{Introduction}

Nuclei instance segmentation, the joint detection, segmentation, and classification of each nucleus, is a
prerequisite for downstream analyses such as tumor grading, immune-infiltration assessment, and prognosis
prediction \cite{nunes2025survey,komura2025ml}. Nuclei are densely packed, morphologically diverse, and show
large staining variation, and a whole-slide image contains tens of thousands of them, so a usable method
must be both accurate and fast.

HoVer-Net \cite{graham2019hover} cast the task as multi-task learning---jointly predicting nuclear pixels
(NP), horizontal-vertical distance maps (HV), and nuclear classification (NC)---and has become the de facto
baseline. The tasks are complementary: the HV maps sharpen the boundary between touching nuclei, the NC labels
keep adjacent nuclei of different classes apart, and all three read the same regions, so one encoder and one
forward pass serve them, and the tasks regularise one another instead of accumulating the errors of a
detect-then-classify pipeline. Its design is nevertheless inefficient: each task gets its own decoder, so three
branches replicate most of the decoder capacity for outputs that differ only in their final channels. Two
lines of work build on this formulation.

Knowledge distillation shrinks a trained model. HoVer-UNet \cite{tommasino2024hoverunet} distils HoVer-Net
into a single-branch U-Net that concatenates the three branch outputs into one tensor, showing that one
decoder can carry all three tasks, and FKDNuSeg \cite{zhao2026fkdnuseg} distils a multi-task nuclei framework
into a 4.11M-parameter student. Both lower inference cost without altering the task formulation, but the student can
only approach its teacher, so distillation presupposes a strong teacher and delivers the accuracy of a model
too large to deploy.

Encoder scaling instead strengthens the encoder, up to pathology foundation models. CellViT
\cite{horst2024cellvit} applied vision transformers (ViTs) with substantial gains but around 700M
parameters, and NuLite \cite{tommasino2026nulite} matches its quality with a FastViT model up to $40\times$
smaller, which places the accuracy in the encoder rather than in the network around it.
Building on SAM \cite{kirillov2023segment} and its medical adaptation MedSAM \cite{ma2024medsam},
CFR-SAM \cite{su2026cfrsam} achieves state-of-the-art results on PanNuke and MoNuSeg, and UNI
\cite{chen2024uni} and Virchow \cite{vorontsov2024virchow} offer general-purpose features from large-scale
pretraining \cite{chanda2024survey}. For most of these models the accuracy is bought with parameter count,
floating-point computation (FLOPs), and GPU memory that clinical workstations and edge devices cannot supply.
Our teacher encoder, UNI2-h, comes from this line.

Neither line makes that accuracy any cheaper. Distillation needs a strong teacher and scaling needs a strong
encoder, so either way the accuracy is produced by the encoder and inherited by the rest of the network. We
therefore keep this accuracy and shrink the decoder, which carries none of it.

We keep the encoder--decoder layout of HoVer-Net but not its three-decoder structure. A single decoder fuses
the encoder features and feeds one lightweight head per task, retaining the capacity of three independent
decoders and matching their accuracy. Because teacher and student share this decoder and differ only in the
encoder, distillation has only the encoder to transfer. NucShare-T pairs UNI2-h with that decoder and reaches
mPQ 0.524 on PanNuke, above every prior method. NucShare therefore contributes \textbf{(1)} a shared decoder
that replaces three task-specific decoders with 37\% of their parameters and no loss in accuracy while keeping
a separate lightweight head per task; \textbf{(2)} output-level distillation to NucShare-S, which keeps the
decoder and borrows an off-the-shelf ConvNeXt V2-Tiny encoder, giving a $20\times$ smaller model that
surpasses CFR-SAM-H on PanNuke mPQ and runs 27.5$\times$ faster per tile than the teacher; and \textbf{(3)} a student that
carries over to an unseen organ without fine-tuning, reaching bPQ 0.668 on zero-shot MoNuSeg against 0.662 for
CFR-SAM-H.

\section{Method}

\begin{figure*}[!t]
\centering
\includegraphics[width=0.95\textwidth]{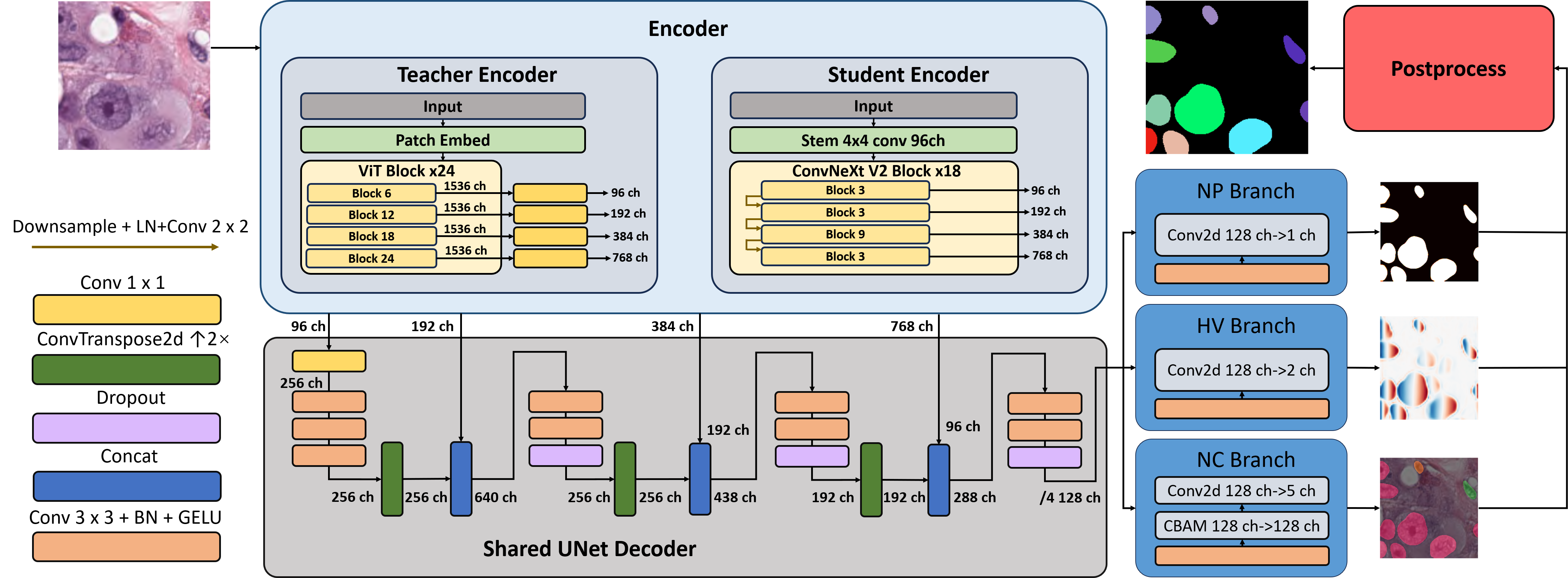}
\caption{Overall architecture of NucShare. The teacher NucShare-T and student NucShare-S are identical
except for the encoder.}
\label{fig:architecture}
\end{figure*}

The overall architecture follows an encoder--decoder pattern, as Fig.~\ref{fig:architecture} shows: an encoder
maps the input to features at four scales, a shared decoder fuses and upsamples them, and three
lightweight heads produce the NP, HV, and NC outputs. The teacher NucShare-T and the student NucShare-S run
the same decoder and differ only in the encoder. We describe the encoders, the shared decoder, and the
distillation objective in turn.

\subsection{Teacher and Student Encoders}

The teacher, NucShare-T, pairs the NucShare decoder with the 683.6M-parameter UNI2-h ViT encoder, taking
features from blocks 6/12/18/24 and projecting each with a $1\times1$ convolution from 1536 to
96/192/384/768 channels, one per decoder level. The student, NucShare-S, keeps the NucShare decoder and borrows an encoder off the shelf: ConvNeXt V2-Tiny
\cite{woo2023convnextv2}, whose Global Response Normalization layers enhance inter-channel feature
competition. Its 27.9M encoder parameters bring the student to 34.7M, about $20\times$ smaller than the
teacher.

\subsection{Multi-Task Shared Decoder}

The NucShare decoder upsamples the four encoder levels to a 128-channel feature map and hands it to the NP,
HV, and NC heads, which follow the multi-task learning paradigm \cite{caruana1997multitask}. Two design
choices follow from what each task needs from the same features.

Upsampling uses learnable transposed convolutions rather than bilinear interpolation. A bilinear kernel is
fixed and smooths uniformly, so it erases exactly the high-frequency gaps between touching nuclei on which NP
boundaries and HV gradients depend, and no single fixed kernel can accommodate three tasks at once.
Transposed convolutions impose no preset smoothing, and because their weights are optimised by the three
losses together, the decoder preserves those gaps instead of erasing them.

The NC head alone carries CBAM attention \cite{woo2018cbam}. Classification needs finer discrimination than
the two geometric tasks, since nuclei of different classes differ only subtly in appearance, and channel and
spatial attention re-weight the shared feature map towards that evidence.

Sharing is more than a parameter-saving device. The three tasks act on the same nuclei at the same
resolution, and all predict pixel-dense maps, so they depend on the same boundary, texture, and region cues
and one decoder serves them without task-specific trunk capacity; because gradients from all three
objectives flow through the same weights, the tasks also regularise one another. Sharing also changes what
distillation must do, since teacher and student run the same decoder and differ only in the encoder: their
outputs are produced by one function in one semantic space, so the distillation error is attributable to the
encoder alone and the student inherits rather than relearns the decoder.

\subsection{Knowledge Distillation}
\label{sec:kd}

Only the encoder is replaced, and the two encoders differ in semantics---we tap four depths of the ViT
and project each to the corresponding width, whereas ConvNeXt yields a native four-level pyramid---so
bridging that gap is what the distillation objective must achieve.

We distill NucShare-T into NucShare-S through output-level knowledge distillation
\cite{hinton2015distilling}. The teacher is trained to convergence with the supervised loss alone and frozen;
the student's NucShare decoder is trained from scratch, and only the student encoder starts from public ConvNeXt V2
weights. The
supervised loss
\begin{equation}\label{eq:sup}
    \mathcal{L}_{\mathrm{sup}} = 2\mathcal{L}_{\mathrm{NP}} + 2\mathcal{L}_{\mathrm{HV}} + 2\mathcal{L}_{\mathrm{NC}},
\end{equation}
combines Focal Tversky $+$ Dice for NP, MSE $+$ MSGE for HV, and Focal $+$ Dice with class-balanced weights
for NC. The student minimizes a convex combination of this loss and the distillation loss:
\begin{equation}\label{eq:total}
    \mathcal{L} = (1-\alpha)\,\mathcal{L}_{\mathrm{sup}} + \alpha\,\mathcal{L}_{\mathrm{KD}},
    \qquad \alpha = 0.5,
\end{equation}
so the supervised term is halved rather than added to in full. The distillation term is itself decomposed
over the three task outputs:
\begin{equation}\label{eq:kd}
    \mathcal{L}_{\mathrm{KD}} = \mathcal{L}^{\mathrm{NP}}_{\mathrm{KL}}
    + \mathcal{L}^{\mathrm{NC}}_{\mathrm{KL}} + \mathcal{L}^{\mathrm{HV}}_{\mathrm{MSE}}.
\end{equation}

\begin{figure}[!t]
\centering
\includegraphics[width=0.80\columnwidth]{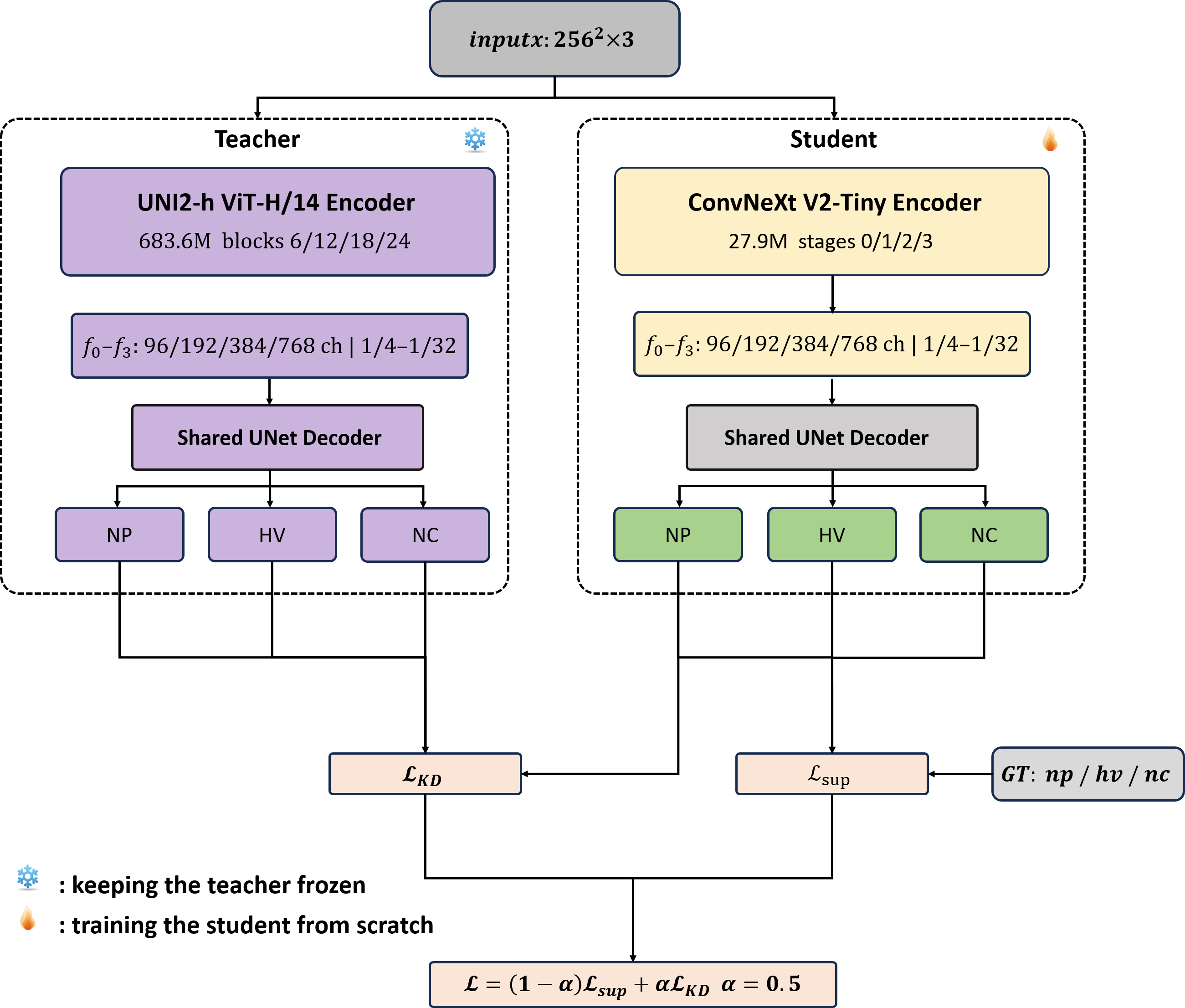}
\caption{Knowledge distillation from NucShare-T to NucShare-S.}
\label{fig:kd}
\end{figure}

The first two terms are Kullback--Leibler divergences over the student and teacher NP foreground
probabilities (sigmoid) and NC class probabilities (softmax); the third is a mean squared error over the
two-channel HV maps, which are continuous and for which a divergence is undefined. All three terms carry
unit weight, with no temperature softening ($T=1.0$), so the student matches the teacher's distributions
directly; temperature applies only to distributions, so the HV branch, a continuous regression target matched
value-wise by MSE, is never softened. Distillation is applied to the three branches independently
(Fig.~\ref{fig:kd}): KL preserves the teacher's uncertainty on ambiguous, densely packed nuclei, whereas MSE
transfers the continuous HV orientation map.

Distillation stays purely output-level: because teacher and student share the decoder, aligning intermediate
features would only re-impose the encoder mismatch the student must absorb.

\section{Experiments}

\subsection{Datasets and Metrics}

PanNuke \cite{gamper2019pannuke} contains $\sim$7,900 $256^2$ H\&E images across 19 tissue types and five
nuclear classes (N/E/I/C/D), split by tissue-level 3-fold cross-validation; the Dead class occupies only
1.53\% of all instances, so we adopt inverse-frequency class-balanced sampling. MoNuSeg
\cite{kumar2017dataset} is used for cross-organ zero-shot evaluation. We report panoptic quality
$PQ = DQ \times SQ$, where $DQ$ is the detection F1 at IoU $>$ 0.5 and $SQ$ the mean IoU of true positives;
mPQ averages $PQ$ uniformly over the five classes and is our primary metric, while binary PQ (bPQ)
aggregates over instances and thus weights frequent classes more heavily. Per-class F1 and detection F1 (Det) measure
classification and instance detection.

\subsection{Comparison with State-of-the-Art}

\begin{figure*}[!t]
\centering
\includegraphics[width=0.68\textwidth]{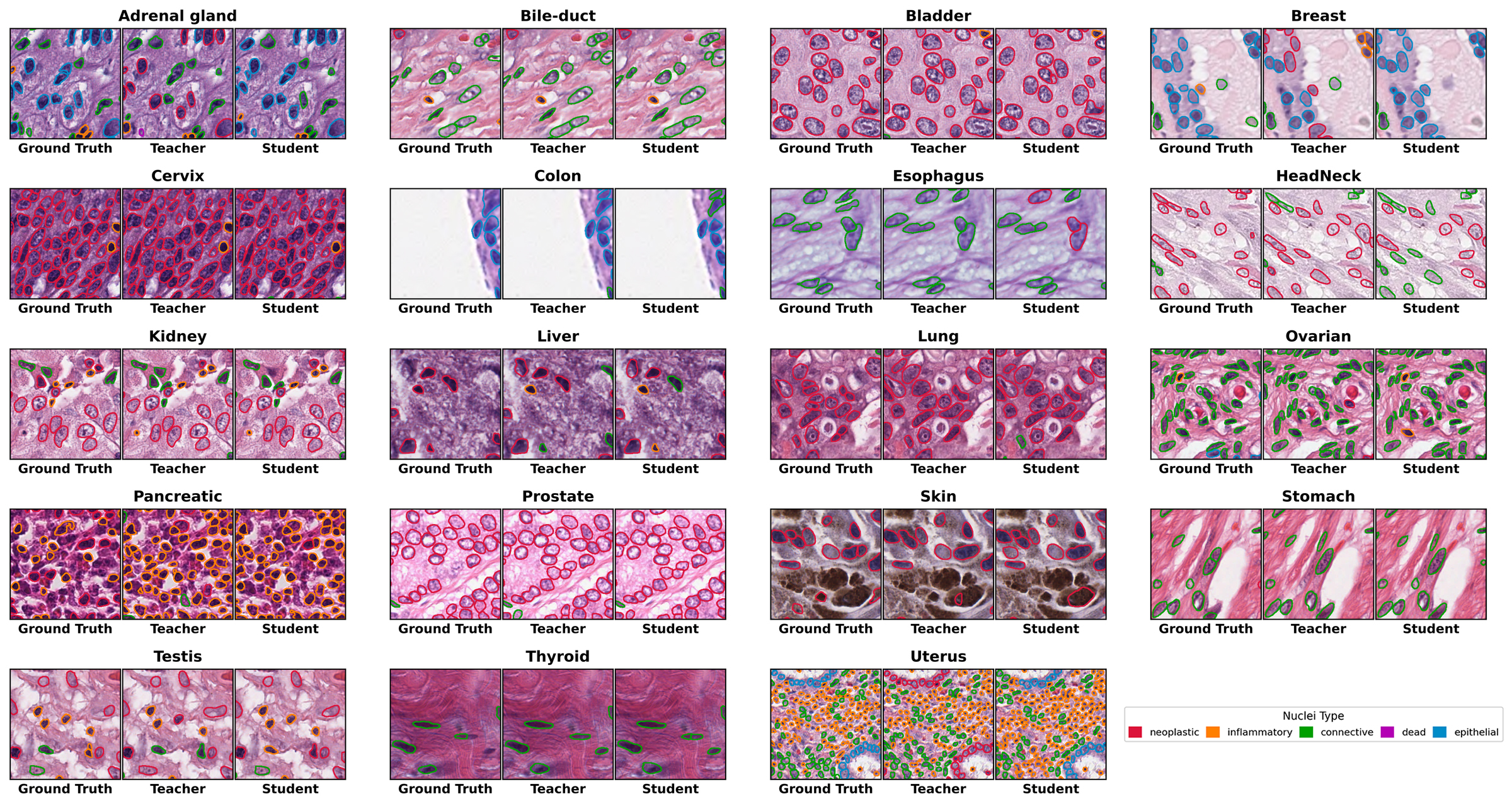}
\caption{Qualitative results on PanNuke. From left to right: Ground Truth, Teacher, Student.}
\label{fig:qual}
\end{figure*}

\begin{table}[t]
\centering
\caption{Comparison of average bPQ and mPQ on PanNuke. Bold indicates the best result and underline the
second best.}
\label{tab:main}
\small
\begin{tabular}{@{}lccc@{}}
\toprule
Model & bPQ & mPQ & Params (M) \\
\midrule
Mask R-CNN \cite{he2017mask} & 0.553 & 0.369 & 41.9 \\
HoVer-Net & 0.660 & 0.463 & 37.6 \\
CellViT-H & 0.679 & 0.498 & 699.7 \\
CFR-SAM-H & \textbf{0.696} & 0.511 & 650.5 \\
\textbf{NucShare-T} & 0.681 & \textbf{0.524} & 690.5 \\
\textbf{NucShare-S} & \underline{0.683} & \underline{0.519} & \textbf{34.7} \\
\bottomrule
\end{tabular}
\end{table}

{\sloppy
Table~\ref{tab:main} reports results averaged over the three folds on PanNuke. NucShare-T attains mPQ 0.524,
above the previous best CFR-SAM-H at 0.511, and NucShare-S retains 0.519 with $20\times$ fewer parameters,
about 1\% below the teacher and about 1/19 the size of CFR-SAM-H. Against FKDNuSeg
\cite{zhao2026fkdnuseg}, NucShare-S improves mPQ by 6.0 points, 0.459 to 0.519, and bPQ by 2.2 points, 0.661 to
0.683. In bPQ,
NucShare-T and NucShare-S, at 0.681 and 0.683, remain below CFR-SAM-H at 0.696. We attribute the bPQ gap to watershed post-processing, which separates densely
overlapping nuclei less robustly than SAM prompting. NucShare-S also leads HoVer-Net and Mask R-CNN
by 5.6 and 15.0 mPQ points.\par}

\begin{table}[t]
\centering
\caption{Per-class F1 on PanNuke. Bold indicates the best result and underline the second best; N/E/I/C/D =
Neoplastic/Epithelial/Inflammatory/Connective/Dead and Det = detection F1.}
\label{tab:f1}
\vspace{2pt}
\small
\begin{tabular}{@{}lcccccc@{}}
\toprule
Model & N & E & I & C & D & Det \\
\midrule
Mask R-CNN & 0.59 & 0.52 & 0.50 & 0.42 & 0.22 & 0.72 \\
HoVer-Net & 0.62 & 0.56 & 0.54 & 0.49 & 0.31 & 0.80 \\
CellViT-H & 0.71 & 0.73 & 0.58 & 0.53 & 0.36 & \textbf{0.83} \\
CFR-SAM-H & \textbf{0.76} & \underline{0.78} & 0.70 & \underline{0.62} & \underline{0.45} & \textbf{0.83} \\
\textbf{NucShare-T} & \textbf{0.76} & 0.74 & \textbf{0.74} & \textbf{0.64} & \textbf{0.49} & \underline{0.81} \\
\textbf{NucShare-S} & \underline{0.75} & \textbf{0.79} & \underline{0.73} & \textbf{0.64} & \textbf{0.49} & \underline{0.81} \\
\bottomrule
\end{tabular}
\end{table}

In the per-class F1 comparison (Table~\ref{tab:f1}), NucShare-S ranks first on Epithelial and ties the best on
Connective and Dead, with 0.79, 0.64, and 0.49, outperforming CFR-SAM-H on Dead by 4 points through
class-balanced sampling and Focal re-weighting; unlike PQ, these scores are unaffected by post-processing.

\begin{table}[t]
\centering
\caption{Zero-shot generalization on MoNuSeg. Bold indicates the best result and underline the second
best.}
\label{tab:monuseg}
\vspace{2pt}
\small
\begin{tabular}{@{}l@{\hspace{4cm}}c@{}}
\toprule
Model & bPQ \\
\midrule
Mask R-CNN & 0.365 \\
HoVer-Net & 0.590 \\
CellViT-H & 0.490 \\
CFR-SAM-H & 0.662 \\
\textbf{NucShare-T} & \underline{0.667} \\
\textbf{NucShare-S} & \textbf{0.668} \\
\bottomrule
\end{tabular}
\end{table}

On zero-shot MoNuSeg (Table~\ref{tab:monuseg}), NucShare-S attains bPQ 0.668, matching the teacher and
exceeding \mbox{CFR-SAM-H} at 0.662, so a $20\times$ reduction in parameters does not hurt cross-organ generalization.
Fig.~\ref{fig:qual} shows qualitative results on PanNuke.

\subsection{Ablation and Distillation Analysis}
\label{sec:ablation}

The decoder-design ablation (Table~\ref{tab:ablation}) shows that the NucShare decoder
matches independent decoders with only 37\% of their parameters, supporting the
implicit regularization of sharing; adding MALA, a multi-scale gated convolution
($3\times3$/$5\times5$/$7\times7$), costs 3.8M parameters and gains nothing. On a single fold, bilinear upsampling and a CBAM-free NC head lose 0.0025 and 0.0022 mPQ.
In distillation, the KD student retains 99.1\% of the teacher's mPQ, against 97.9\% when
trained from scratch. Adding MALA to the student's decoder lowers mPQ to 0.5175, and an MSE alignment term
($\lambda{=}0.1$) on the encoder features $f_0$--$f_3$ lowers it to 0.5150.

\begin{table}[b]
\centering
\caption{Decoder-design ablation on PanNuke.}
\label{tab:ablation}
\vspace{2pt}
\resizebox{\columnwidth}{!}{%
\begin{tabular}{@{}lccc@{}}
\toprule
Configuration & Decoder Params (M) & mPQ & bPQ \\
\midrule
NucShare decoder (baseline) & 6.9 & 0.5238$\pm$0.0093 & 0.6815$\pm$0.0082 \\
$+$MALA & 10.7 & 0.5237$\pm$0.0093 & 0.6811$\pm$0.0073 \\
UNet3 (independent) & 18.8 & 0.5236$\pm$0.0113 & 0.6803$\pm$0.0058 \\
\bottomrule
\end{tabular}}
\end{table}

\subsection{Parameters and Inference Speed}

NucShare-S reduces the teacher from 690.5M to 34.7M parameters and from roughly 200 to 11.6 GFLOPs per
tile; at batch size 1 it occupies 0.2GB of GPU memory against 2.8GB, so inference fits on a 4GB card. On an
A100 with $256^2$ tiles, it reaches 708.7 patches/s against 25.8, 27.5$\times$ faster, at 1.4 ms per tile.
End-to-end on a $1024^2$ image, NucShare-S takes 0.14 s against 1.03 s (7.4$\times$); the smaller margin
reflects tiling and post-processing overhead.

\section{Conclusion}

We presented NucShare, a lightweight model whose shared decoder matches three independent decoders at
37\% of their parameters and lets a student differing only in its encoder be distilled from a UNI2-h teacher.
NucShare-T reaches mPQ 0.524 on PanNuke, above CFR-SAM-H at 0.511, and NucShare-S retains 99.1\% of that at
34.7M parameters, runs 27.5$\times$ faster per tile than the teacher, needs 0.2GB of GPU memory, and attains bPQ
0.668 on zero-shot MoNuSeg against 0.662 for CFR-SAM-H. On PanNuke, bPQ still trails CFR-SAM-H, which we trace
to watershed post-processing, so improving it together with INT8 quantization and TensorRT acceleration is
left for future work.

\newpage
\begin{center}{\bf COMPLIANCE WITH ETHICAL STANDARDS}\end{center}
This research study was conducted retrospectively using human subject data from the open-access PanNuke and
MoNuSeg datasets. Ethical approval was not required as confirmed by the license attached with the open access
data.

\begin{center}{\bf ACKNOWLEDGMENT}\end{center}
No funding was received for this work. The authors have no conflicts of interest to declare. The code is
available at \url{https://github.com/L-iMeng-L/Newidea}; trained weights are available from the authors on
request.

\end{document}